# TGGLinesPlus: A robust topological graph-guided computer vision algorithm for line detection from images

Liping Yang [1,2,3 *], Joshua Driscol [1,2,4 ‡], Ming Gong [5‡], Katie Slack [1,2], Wenbin Zhang [6], Shujie Wang [7,8], Catherine G. Potts [9]

*corresponding author lipingyang@unm.edu
‡ equal contribution

**Authors institutional affiliations:**

1. GeoAIR Lab, Department of Geography and Environmental Studies, University of New Mexico, Albuquerque, NM 87131, USA
2. Center for the Advancement of Spatial Informatics Research and Education (ASPIRE), University of New Mexico, Albuquerque, NM 87131, USA
3. Department of Computer Science, University of New Mexico, Albuquerque, NM 87106, USA
4. Woodwell Climate Research Center. Falmouth, MA 02540, USA
5. Department of Electrical and Computer Engineering, University of Dayton, Dayton, OH, 45469, USA
6. Knight Foundation School of Computing and Information Sciences, Florida International University, Miami, FL 33199, USA
7. Department of Geography, The Pennsylvania State University, University Park, PA 16802, USA
8. Earth and Environmental Systems Institute, The Pennsylvania State University, University Park, PA 16802, USA
9. D-Wave Quantum Inc., Burnaby, British Columbia, Canada

### Abstract

Line detection is a classic and essential problem in image processing, computer vision and machine intelligence. Line detection has many important applications, including image vectorization (e.g., document recognition and art design), indoor mapping, and important societal challenges (e.g., sea ice fracture line extraction from satellite imagery). Many line detection algorithms and methods have been developed, but robust and intuitive methods are still lacking. In this paper, we proposed and implemented a topological graph-guided algorithm, named *TGGLinesPlus*, for line detection. Our experiments on images from a wide range of domains have demonstrated the flexibility of our *TGGLinesPlus* algorithm. We benchmarked our algorithm with five classic and state-of-the-art line detection methods and evaluated the benchmark results qualitatively and quantitatively, the results demonstrate the robustness of *TGGLinesPlus*.

### Abbreviations

The following abbreviations (ordered alphabetically) are used in this article:

CV
: Computer Vision

CT scan
: Computed Tomography scan

EDLines
: Edge Drawing Lines

GIS
: Geographic Information Systems

HT
: Hough transform

LSD
: Line Segment Detector

MNDWI
: Modified Normalized Difference Water Index

NDWI
: Normalized Difference Water Index

OCR
: Optical Character Recognition

OLI
: Operational Land Imager

PPHT
: Progressive Probabilistic Hough Transform

RS
: Remote Sensing

TGGLines
: Topological Graph Guided Lines

*TGGLinesPlus*
: Topological Graph Guided Lines Plus

YAML
: *yet another markup language* (initial acronym) or *YAML ain't markup language* (it was repurposed as *YAML Ain't Markup Language*, a recursive acronym, to distinguish its purpose as data-oriented, rather than document markup)

## 1. INTRODUCTION AND MOTIVATION

Line detection has been studied for decades in computer vision and image processing and plays a fundamental and impactful role for advanced vision-based applications ranging from medical image vectorization for advanced analysis (Ricci & Perfetti, 2007), automatic linear feature extraction from satellite images to detect features (Tupin et al., 1998), to lane detection for autonomous vehicles (Lai et al., 2020), and indoor mapping (An et al., 2012). Many line segment detection methods are developed for natural images (Akinlar & Topal, 2011; Ballard, 1981; Cho et al., 2018; Grompone von Gioi et al., 2008; Matas et al., 2000). However, those methods do not work well for computer generated, scanned document images, and sensor generated images (e.g., satellite imagery). Solutions to those challenging types of images can also be helpful for document image analysis tasks such as optical character recognition (OCR) (Yang et al., 2020).

We present a robust and topological graph-guided approach for line segment detection in images. Our method mimics how humans perceive the world (using the concept of invariance in topology) (Chen, 1982; Tozzi & Peters, 2019) and provides an effective topological graph-based image representation in computer vision to tackle challenges in image analysis. Our method also introduces a novel and robust technical approach through the integration of computational geometry and topological relations among objects present in an image. Specifically, our approach combines the power of topological graphs (Yang & Worboys, 2015) and image skeletons (Yang et al., 2019a, 2019b) to generate a skeleton graph representation to tackle line segment detection challenges. A skeleton is a central line (1-pixel wide) representation of an object present in an image obtained via thinning (Lam et al., 1992; Yang et al., 2019a). The skeleton emphasizes topological and geometrical properties of shapes (Yang et al., 2019a). In our approach, *the image skeleton serves as the essential bridge from pixel image representation to topological graph representation*. We compare our approach with five state-of-the-art line segment detection methods. The benchmark results demonstrate that our approach outperforms other methods both qualitatively (visually) and quantitatively. Our method has two main advantages. **(1)** As our method detects line segments and organizes the segments into a topological graph, we return to users a dictionary of useful information (detailed in Algorithm 1 in Section 3.4.1 and further detailed in Step 5 in Section A in the supplementary materials (see **DATA AND CODES AVAILABILITY STATEMENT**). The resulting graph and all useful information in the returning dictionary can be used by advanced machine vision and machine learning tasks, such as shape analysis, line segment-based image retrieval, feature extraction, and character recognition. **(2)** Our *TGGLinesPlus* algorithm does not require any parameters and thus there is no need to tune parameters for best results.

Line detection performance is often simply evaluated by visualization (i.e., visual checks), which is a non-quantitative evaluation. In this paper, our *TGGLinesPlus* algorithm consistently outperforms other state-of-the-art line segment detection methods, including our own previous TGGLine algorithm. The strengths of our *TGGLinesPlus* algorithms over TGGLines and other state-of-art methods are detailed in Section 5.2. One very noticeable novel contribution of our work in this paper is our evaluation of the results both qualitatively (visually, see Section 4.2) and quantitatively (Section 4.3). One of the most difficult problems for quantitative evaluation is manually annotating line segments in an image accurately. Sometimes even humans will annotate lines differently due to the zigzag "noise" introduced by the scanning process (see the annotation examples provided in Figure 8) and the subjective decisions of the human annotator. Typically for a scientific diagram image, there can be several hundreds of lines that must be annotated, see image #09 (contour map) in Figure 6 and image #10 (document page) in Figure 7 for examples. We provide a simple interface for line segment annotation to generate ground truth data, see the ground truth (GT) column for each benchmark image shown in Figure 4 to Figure 7, as well as quantifiable metrics using graph properties from graph theory, to evaluate line detection performance, detailed in Section 4.3.

The rest of the paper is structured as the following. Section 2 covers related work, including existing line segment detection methods and the topological graph-based image representations that our method is built on. Section 3 focuses on our proposed approach and algorithms: a brief introduction to the image representation (Section 3.1) we use and the main graph theory concepts (Section 3.2) involved in our approach, followed by the node types (Section 3.3) and the *TGGLinesPlus* algorithm and workflow illustration (Section 3.4), visually illustrated by two simple and straightforward examples. In Section 4, we present our experiment setup (Section 4.1), and the benchmark results comparing five state-of-the-art methods (Section 4.2), followed by the quantitative evaluation (Section 4.3). In Section 5, we provide a discussion, describe strengths (Section 5.2), point out limitations and future work (Section 5.3), and discuss potential geospatial and general applications (Section 5.4). The paper concludes in Section 6 with a mention of potential applications. For readability, we provide a list of abbreviations right below the abstract.

## 2. BACKGROUND

We here discuss some state-of-the-art line segment detection methods and the topological-based image representations and approaches that our method is built upon. Recent reviews of line segment detection methods can be found in (Guerreiro & Aguiar, (2012) and Rahmdel et al. (2015). State-of-the-art line segment detection methods are either edge-based or local gradient-based. Edge-based line detectors include the Hough transform (HT) (Ballard, 1981; Rahmdel et al., 2015), which is one of the most well-known line detection methods. The HT is a feature extraction technique for detecting simple shapes such as lines and circles in an image. As the HT is based on edge detection, the quality of detected lines depends heavily on the quality of the edge map, which often uses a Canny edge detector (Canny, 1986) to pre-process images. From an edge map, the HT searches possible line configurations through voting on the contribution of each edge pixel, after which lines are detected based on a threshold (the minimum vote for whether a line should be considered as a line). The main drawback of the HT is that it is computationally expensive, and it only outputs the parameters of a line equation, not the endpoints of line segments. The progressive probabilistic Hough transform (PPHT) (Matas et al., 2000) is an optimization of the standard HT; it does not take all the points into consideration, instead taking only a random subset of points that are sufficient for line detection. The PPHT detects the two endpoints of each line and can be used to detect line segments present in images.

Local gradient-based line detectors are successful on natural images, but not on images such as document characters (See Section 4.2). Unlike the HT and PPHT introduced above, the line segment detector (LSD) (Grompone von Gioi et al., 2008), another well-known line segment detection technique, is a local gradient-based method as it detects line segments locally using gradient orientation values. LSD is designed to work on any digital image without parameter tuning. LSD has been tested on a wide set of natural images and can often generate good results. A linear time edge drawing line segment detection method, called EDLines, was proposed to speed up LSD. EDLines used the concept of edge drawing to produce an edge map; the main idea is to use image gradient information to connect the edge pixels that belong to the same segment (Akinlar & Topal, 2011). Like LSD, EDlines is also based on local gradients and it requires no parameter tuning; however, it is not very robust (see benchmark in Figure 4 to Figure 7 in Section 4.2). A recent line segment detection method called linelet (Cho et al., 2018) uses linelets to represent intrinsic properties of line segments in images using an undirected graph. The major steps include detecting linelets and grouping linelets. Their results show that the linelet method performs well on an urban scene dataset, but it does not work well for other types of images (see Section 4.2, in particular in Figure 4 to Figure 7).

Our approach, on the other hand, is inspired by and makes use of a topological graph-based image representation, called a skeleton graph, initially proposed in (Yang et al., 2019b). First, we use the well-known and robust Zhang-Suen thinning algorithm (T. Y. Zhang & Suen, 1984) to extract image skeletons out of images from various domains. An image skeleton is a central line (1-pixel wide) representation of an object present in an image obtained via thinning (Lam et al., 1992; Yang et al., 2019a) that captures and emphasizes the topological and geometrical properties of shapes making up objects present in an image (Komala Lakshmi & Punithavalli, 2009; Yang et al., 2019a). Using the image skeleton, we automatically generated the skeleton graph, which is then used to automatically extract topological features that can be fed into different machine learning algorithms for image classification (Yang et al., 2019b). To our knowledge, our previous work TGGLines (Gong et al., 2020) is the first important attempt taking advantage of graph theory to tackle computer vision problems for line detection. However, TGGLines is not computationally efficient (at least its implementation is very slow, see benchmark runtimes in Section B.2 in the supplementary materials (see **DATA AND CODES AVAILABILITY STATEMENT)**, and its performance significantly suffers when artifacts introduced (e.g., “fake” turns in lines) (see benchmark results, Figures 4 to 7 in Section 4.2; detailed in the discussions in Section 5.1).

## 3. APPROACH AND ALGORITHMS

We now elaborate on our method, which includes image representation (Section 3.1), graph theory concepts (Section 3.2), node types (Section 3.3), and workflow illustrated using two straightforward examples (Section 3.4), followed by algorithms (Section 3.5).

### 3.1 Image representation

The image representation used in our approach is the skeleton graph initially proposed in (Yang et al., 2019b), which is a topological graph generated from the skeleton of an image. We use the Zhang-Suen thinning algorithm (T. Y. Zhang & Suen, 1984) for image skeleton extraction as it is well-known and robust. See Figure 1 below for an illustration of the image representation we use in our approach (the handwritten digit image used here is taken from the MNIST dataset (Lecun et al., 1998)). In the skeleton graph, each node represents a pixel in the image skeleton, and edges are used to connect pairs of nodes that are neighbors in the image skeleton.

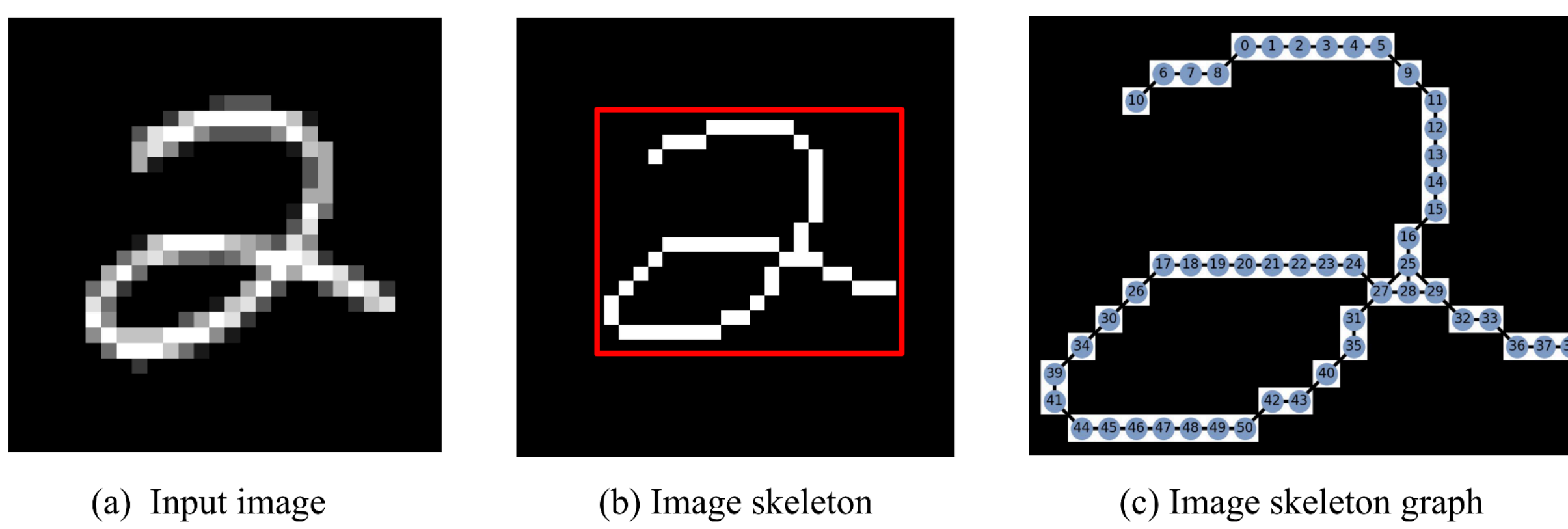

(a) Input image (b) Image skeleton (c) Image skeleton graph

Figure 1. An example of the skeleton graph image representation. (a) is the input image. (b) shows the image skeleton extracted from the input image. (c) provides the skeleton graph generated from the image skeleton present in (b). In the image skeleton graph, each node represents a pixel in the image skeleton, and each edge indicates that the two pixels it connects are neighbors in the image skeleton. Note that the image in (c) is cropped to improve readability.

### 3.2 Graph theory concepts

Here we provide a brief introduction to some key graph theory concepts that our *TGGLinesPlus* algorithm builds upon, to get our readers prepared to understand the algorithms in Section 3.5. The sources we referenced for the concepts definitions include *Bondy & Murty (2008), Skiena*

*(1991), Trudeau (2013), Wilson (1979), Yang (2015), Yang et al., (2019b), Yang & Worboys (2015), P. Zhang & Chartrand (2006).*

**Graph:** In graph theory, a graph *G=(N, E)* is a mathematical data structure consisting of two sets *N (G)* and *E (G)*. *N* is a non-empty finite set, and the elements of *N* are called nodes (also called vertices). *E* is a finite set, and the elements of *E* are called edges; more specifically, *E* includes unordered/ordered pairs of elements of *N*. Note that *E* can be empty, in which case *G* is called a null graph. A graph *G* can be directed or undirected. In this paper, we deal with undirected graphs, which means *E* consists of unordered pairs of elements of *N*.

**Subgraph:** a subgraph of a graph *G* is a graph whose nodes belong to *N(G)* and each of whose edges belongs to *E(G)*. A non-null graph can have many subgraphs, but in this paper, we leverage connected components to decide the number of subgraphs. For example, if a graph only contains one connected component, we treat the graph as the single subgraph. If a graph contains two connected components, we treat each of the connected components as a subgraph.

**Path:** A path in a graph *G=(N, E)* is a simple linear subgraph of *G*. More specifically, a subgraph *G'= (N',E')* such that $N' \subseteq N$, $E' \subseteq E$, and every node has exactly two edges. We can understand a path as a subgraph where the first and the last nodes have a degree of one, and the other nodes have a degree of two. The degree of a node refers to the number of edges connected to the node. From a less technical perspective, a path is a sequence of non-repeated nodes connected through edges present in a graph. A path is basically a connected line in a graph; we could draw it without lifting a pencil.

**Connected component:** A connected component of an undirected graph *G* is a maximal connected subgraph of *G.* Two nodes are in the same connected component if and only if there exists a path between them. A graph is connected if and only if it has exactly one connected component.

**Image skeleton graph:** An image skeleton graph is an embedded graph *G* generated from an image skeleton, where each node represents a pixel in the image skeleton, and an edge between two pixel nodes indicate the two pixels are neighbors.

**Complete graph:** A complete graph *G* is an undirected graph in which every pair of distinct vertices is connected by a unique edge**.**

**Clique:** A clique of an undirected graph *G* is a set of nodes such that every two distinct nodes in the clique are adjacent, implying that a clique of a graph *G* is a complete subgraph of *G*. In this paper, we calculate *a clique of size 3* to detect primary junction node(s) from an initial set of junction nodes (see Section 3.3 below).

**Cycle:** In graph theory, a cycle consists of a sequence of adjacent and distinct nodes in a graph. The only exception is that the first and last nodes of the cycle sequence must be the same node. In other words, a path that starts from a given node and ends at the same node is called a cycle.

### 3.3 Node types in image skeleton graph

There are two types of *path segmentation endpoints* in our *TGGLinesPlus*. We will use them to segment a simplified image skeleton graph to multiple paths: (1) **Terminal node:** A (pixel) node that has only 1 neighbor. (2) **Junction node:** A (pixel) node that has *n* neighbors where $n > 2$.

Note that in our algorithm, we first compute the initial set of junction nodes and then we locate primary junction nodes. Primary junction nodes could be the original junction nodes or a subset of the original junction nodes and are used with the list of terminal nodes to segment paths in a given graph.

### 3.4 *TGGLinesPlus* algorithm and workflow illustrations

In this section, we provide our *TGGLinesPlus* algorithm and illustrate the algorithm workflow with two simple examples.

#### 3.4.1 Algorithms

In this section, we provide the algorithms we developed for *TGGLinesPlus*. The pseudo code for the *TGGLinesPlus* algorithm is provided in Algorithm 1; whereas Algorithm 2 and Algorithm 3 are the sub-algorithms including more details for steps 3.1 and 3.2 in Algorithm 1. For a given input image, we first compute its binarization and then image skeleton, using the image binary as input. The resulting image skeleton is then fed into the main *TGGLinesPlus* algorithm. A skeleton graph is then generated from the image skeleton. The skeleton graph is then split into subgraphs. For each subgraph, a set of junction nodes and terminal nodes (defined in Section. 3.3) are detected by counting the number of incident edges *n* for each pixel node in the skeleton graph. (Two edges are incident if they are connected by one shared node.) We then leverage cliques to detect primary junction nodes from a set of initial junction nodes (detailed in Section 3.3 above), and then to identify which edges might "unnecessarily" complicate path segmentation and thus can be removed to simplify the original image skelton graph and get ready for path segmentation. This is important as this will reduce the number of tiny paths that complicate our segmentation algorithm. Note that we exclude these removed edges as paths in the algorithm, but return them back to the user in the output dictionary in Algorithm 1 if those edges identified to be removed are important for users' domain problems. A more detailed elaboration on each major step with more implementation specifications is provided in Section A in the supplementary materials (see **DATA AND CODES AVAILABILITY STATEMENT**), which will serve as an important guide for those who would like to use the algorithm and/or further improve the algorithm from our open-sourced *TGGLinesPlus* algorithm Python code (see **DATA AND CODES AVAILABILITY STATEMENT** for the GeoAIR Lab GitHub repository link to our algorithm).

---

**Algorithm 1: Our proposed *TGGLinesPlus* algorithm** *(see detailed elaboration of the algorithm with implementation specifics for each algorithm step in Section A in the supplementary materials (see **DATA AND CODES AVAILABILITY STATEMENT**)*

---

**Input:** An image *I*
**Output:** A dictionary *d*
0.1: B ⟵ binarization (*I*) //pre-processing to create a binarized image from input image
0.2: *S* ⟵ skeletonization (*B*) //create an image skeleton from a binarized image
1: *G* ⟵ skeleonGraph (*S*) //generate an image skeleton graph from an image skeleton
2: *Lg* ⟵ subgraph (*G*) //split the main graph into a list of subgraphs
3: **for each** subgraph *g* in *Lg* **do:**
3.1 **simplify** current subgraph *g* // detailed in **Algorithm 2** below.
3.2 **segment** the simplified graph *g* into a list of paths *Lp* //detailed in **Algorithm 3** below
4: **merge** the lists that each *Lp* stores for the segmented paths of a simplified graph *g*

5: **return** a dictionary *d that* stores useful information

**Notes:** The algorithm outputs information that may be helpful for users across several disciplines, such as lists of primary junction nodes, terminal nodes, and cliques. More details are available in Section A in the supplementary materials (see **DATA AND CODES AVAILABILITY STATEMENT**).

---

---

**Algorithm 2:** Simply a graph by detecting primary junction nodes and removing unnecessary edges

---

**Input:** A subgraph *g* //see line 3 and line 3.1 in Algorithm 1 above
**Output:** A simplified subgraph *g*
1: find cliques from junctions subgraphs //see clique definition in Section 3.2
2: remove edges //simplify the graph
3: identify the path segmentation endpoints for the subgraph //see Section 3.3 for two types of path segmentation endpoints

**Notes:** Fuller implementation specifics, for example, how we identify the edges to be removed using cliques are detailed in the Step 3.1.1 to 3.1.3 in Section A in the supplementary materials (see **DATA AND CODES AVAILABILITY STATEMENT**).

---

---

**Algorithm 3:** Segment a simplified graph *g* into paths

---

**Input:** A simplified subgraph g //see line 3, 3.1, and 3. 2 in Algorithm 1 above
**Output:** A list *Lp* storing the segmented paths of the input simplified graph *g*
1: create a copy of the simplified graph for path segmentation
2: compute initial paths list *Lp* //path definition in graph theory are detailed in Section 3.2
3: check for cycles in the graph //cycle definition in graph theory are detailed in Section 3.2
4: split initial paths list *Lp* into lists containing only the two *path segmentation endpoints* //see Section 3.3. for two types of path segmentation endpoints

**Notes:** Fuller implementation specifics, for example, why we make a copy of the simplified graph for path segmentation are detailed in the Step 3.2.1 to 3.2.4 in Section A in the supplementary materials (see **DATA AND CODES AVAILABILITY STATEMENT**).

---

### 3.4.2 Workflow illustrations

The workflow of our approach is illustrated visually by two simple and straightforward examples, presented in Figure 2 (a single graph) and Figure 3 (a graph containing two subgraphs). Let us start with the illustration using the simple example shown in Figure 2. Given an input image shown in step 1 in Figure 2, the image skeleton (step 3 in Figure 2) is extracted from the binarized image (step 2 in Figure 2) of the input image. Then an image skeleton graph (step 4 in Figure 2) is generated from the image skeleton. In the next step, the initial set of junction nodes (detailed in Section 3.3) are used to compute cliques of size 3 (step 5 in Figure 2).

*For example, in step 5 in Figure 2 there are 4 initial junction nodes, and it forms 2 cliques of size 3. We use the cliques of the initial set of junction nodes to decide which edges need to be removed.* Next, we identify which edges to be removed (see step 6 in Figure 2); the two red edges are marked to be removed to simplify the original image skeleton graph. Now we get the simplified graph (see step 7 in Figure 2) and then we get a set of primary junction nodes (see the two red nodes 27 and 28 in step 8 in Figure 2) for an example. Primary junction nodes and terminal nodes (detailed in Section 3.3, see step 9 in Figure 2) are used to segment the simplified graph (see step 7 in Figure 2). Finally, we get the three segmented paths (seen in step 10 in Figure 2): **(1)** the path from the terminal node 10 to the primary junction node 28, **(2)** the path from the terminal node 38 to the primary junction node 28, and **(3)** a loop path starting and ending at the primary junction node 27.

For the subgraph examples illustrated in Figure 3, they have the exact same steps as the example shown in Figure 2; the only difference is that it computes the connected components for subgraphs (see step 5 in Figure 3) after we get the image skeleton graph (see step 4 in Figure 3). We see from steps 6 to 10 in Figure 3 that there are two subgraphs doing the same process in sequence as steps 5 to 10 in Figure 2, then we need to merge the segmented paths from the two subgraphs (see step 11 in Figure 3).

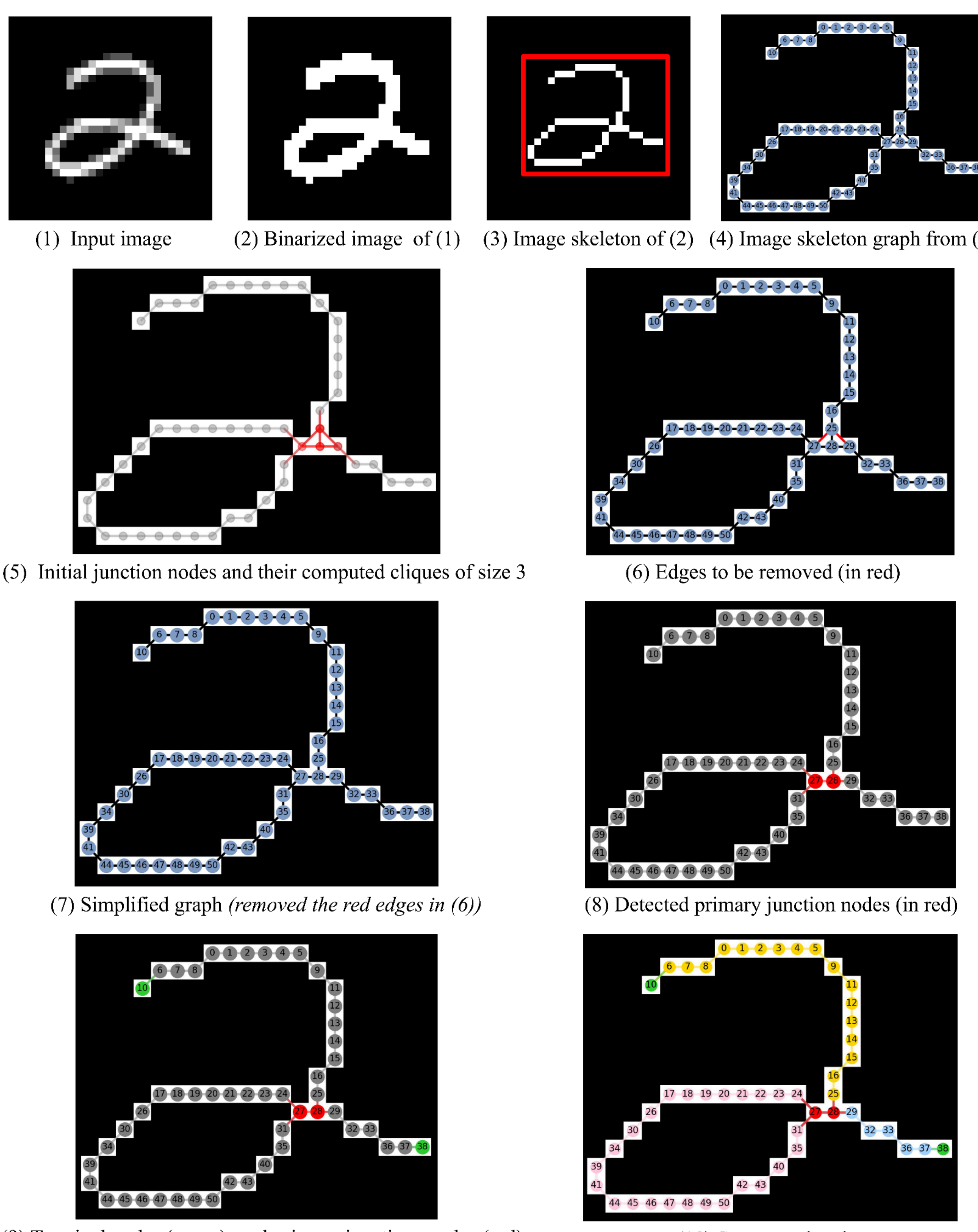

(1) Input image (2) Binarized image of (1) (3) Image skeleton of (2) (4) Image skeleton graph from (3)

(5) Initial junction nodes and their computed cliques of size 3 (6) Edges to be removed (in red)

(7) Simplified graph *(removed the red edges in (6))* (8) Detected primary junction nodes (in red)

(9) Terminal nodes (green) and primary junctions nodes (red) (10) Segmented paths

Figure 2. An illustration of the *TGGLinesPlus* workflow (a single connected component graph example). Step 5 are cliques from the image skeleton graph in step 4. Note that the image in step 4 (and all later steps) has cropped the black background to improve readability. In step 9, terminal nodes (green) and primary junction nodes (red) show the path segmentation endpoints we use to segment paths. Note that the nodes in step 5 will usually have labels, we remove the node labels for readability in the figures.

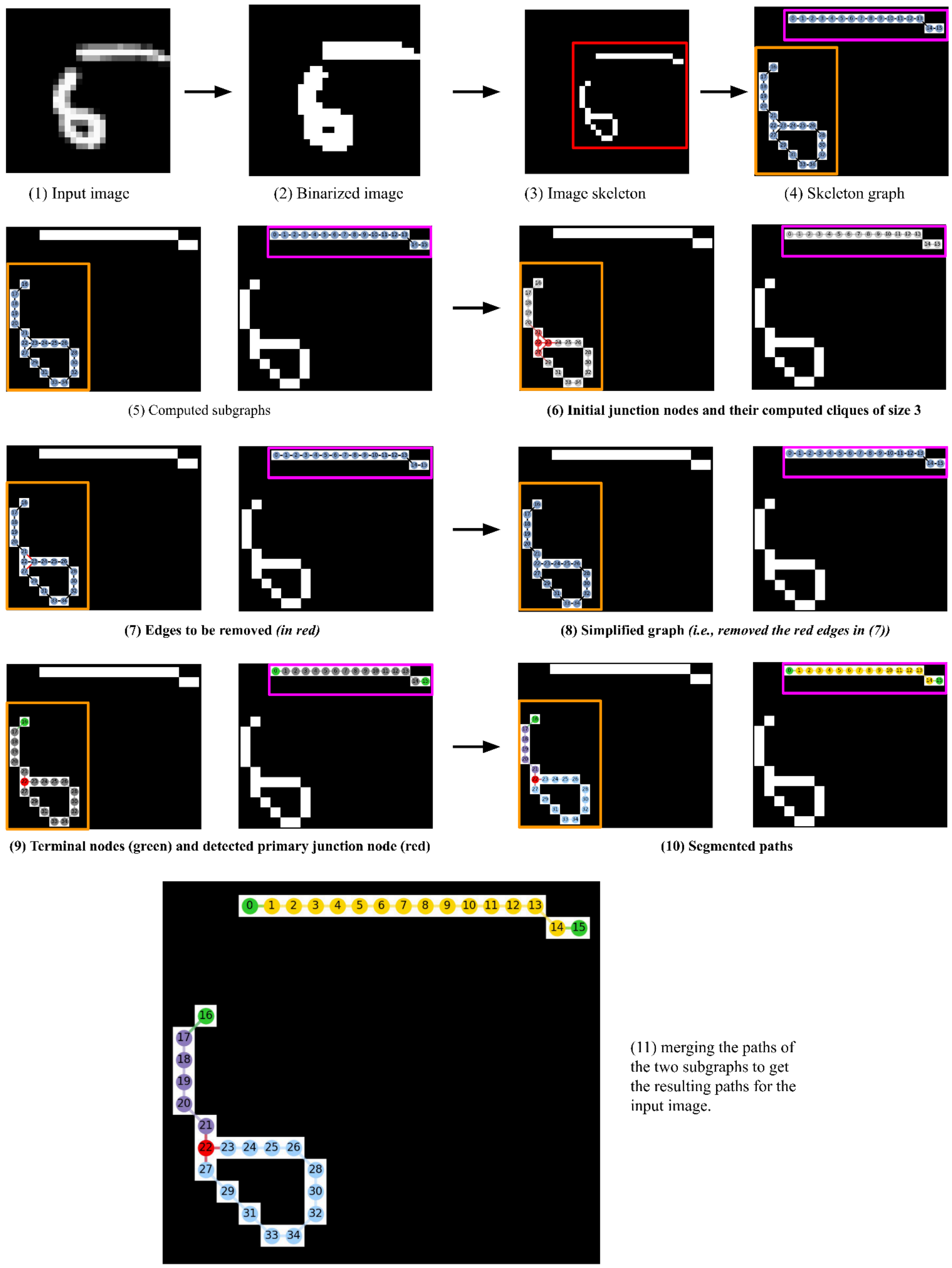

Figure 3. An illustration of the *TGGLinesPlus* workflow (a two-subgraph example). Note that the steps in bold (steps 6 to 10) are illustrating the same process in sequence for the orange subgraph and the pink subgraph. We segment paths in each subgraph sequentially (not in parallel): the top flat subgraph first, and then the subgraph with the loop second. To save space we include steps 6 to 10 side by side. Once the paths for each subgraph are computed, they are merged in step 11. The red box in step 3 indicates a cropped and zoomed-in step 4 for readability. The top right subgraph indicated by the pink box has one path. The bottom left subgraph illustrated by the orange box has two segmented paths: one is from terminal node 16 to the primary junction node 22 in step 11, and the other is a loop path starting and ending at the primary junction node 22. So the whole input image has three segmented paths.

#### 3.4.3 Exemplary dictionary of *TGGLinesPlus* algorithm output

We demonstrate the usefulness of the output from our proposed and implemented *TGGLinesPlus* algorithm. A detailed explanation of the results dictionary of the two illustrated examples, the digit in Figure 2 and the digit in Figure 3 above, can be found in our open-sourced repository. We have one specific jupyter notebook file for this purpose; see **DATA AND CODES AVAILABILITY STATEMENT**.

## 4. EXPERIMENTS, BENCHMARKING AND RESULTS

### 4.1 Experiments setup

Our *TGGLinesPlus* algorithm was implemented in Python. In our open-source GitHub repository (See ***DATA AND CODES AVAILABILITY STATEMENT***), we have created a YML file so that coding environment dependencies can be easily installed and our results can be quickly replicated. More detailed specifications about the computing environment for the benchmark are provided in our open-sourced GitHub repository, check Section B in the supplementary materials (see **DATA AND CODES AVAILABILITY STATEMENT**).

### 4.2 Experiments and benchmark on images from various domains

We ran experiments of our algorithm *TGGLinesPlus* on images from diverse domains (character recognition/document digitization, health and biology images, earth sciences satellite imagery of a sea ice shelf in Antarctica, a scanned contours map, transportation and route planning, structural engineering concrete cracks images) and under various scenarios (small vs. large image size, simple vs. complex examples, single component graph vs. graphs with many subgraphs; graphs with/without cycles). Details about the data sources are provided in Section B.1 in the supplementary materials (see **DATA AND CODES AVAILABILITY STATEMENT**) (particularly Table 1). We also benchmarked *TGGLinesPlus* with five classic / state-of-the-art line detection methods (see Section 2 for details for the five methods): PPHT (Matas et al., 2000), LSD (Grompone von Gioi et al., 2008), EDLines (Akinlar & Topal, 2011), Linelet (Cho et al., 2018), and TGGLines (Gong et al., 2020). The benchmark results of the five methods and our new *TGGLinesPlus* algorithm are provided in Figure 4 to Figure 7. The details about the benchmark data sources are provided in Section B.1 in the supplementary materials (see **DATA AND CODES AVAILABILITY STATEMENT**); the computation configuration, benchmark runtime, and benchmark implementation specifications for each of the benchmark methods, including our *TGGLinesPlus* are detailed in in Section B.2 in the supplementary materials (see **DATA AND CODES AVAILABILITY STATEMENT**).

From benchmark images in Figure 4, PPHT, LSD, EDLines, Linelet, and our previous TGGLines algorithm do not perform well for small images. Several of the algorithms (PPHT, EDLines, Linelet) produced no output at all for the MNIST and CMNIST images we analyzed. The LSD and TGGLines algorithms were able to generate outputs, but produced outlines that poorly reproduced the input images. *TGGLinesPlus*, though, works well at the individual pixel level because its graph representation creates a node for every pixel in the image skeleton. This allows *TGGLinesPlus* to follow contours and paths exactly as they are instead of relying on an algorithm where tunable parameters are necessary to generate output lines.

Images in benchmark Figures 5 (road, retina, and cement cracks), and Figure 6 (remote sensing Landsat 8 OLI, and contours map) are much larger and more complex than the images in Figure 4 (MNIST and CMNIST images) and the original input image is already fairly close to an image binary or skeleton. In this case, the PPHT, LSD, and Linelet algorithms perform reasonably well and do much better than they did for small images in Figure 4 in our analysis. However, Linelet, much like the EDLines algorithm, suffers from issues related to discontinuity between lines or poor/missing line recognition for large parts of the image. TGGLines performs better than PPHT, LSD, EDLines, and Linelet by offering full coverage of the lines in an image skeleton (no breaks, more realistic lines). While the original TGGLines method works well, though, *TGGLinesPlus* offers an improvement in two key areas, apart from working better on small images, discussed above for images (image #01 to #03) in benchmark Figures 4. **1)** Paths in the resulting vector representation of the image are more intuitive and continuous. TGGLines used turning nodes (a bend in a path composed of 2 nodes that change direction) which separated lines that should have been the same path. *TGGLinesPlus* does not use turning nodes and is able to maintain paths over curved surfaces without breaking the path into multiple line segments. It does so by using a graph to follow the image skeleton exactly and only separating distinct paths using junction and terminal nodes (Section 3.3). **2)** *TGGLinesPlus* returns the starting and ending nodes for every path in the resulting vector representation of the image. We believe this will help researchers analyze graph path statistics and also to combine different paths as "routes" if they choose to do so. More detailed strengths of *TGGLinesPlus* over TGGLines and other benchmark methods are provided in Section 5.2.

In Figure 7, only TGGLines and *TGGLinesPlus* are able to return continuous paths that represent the input image. This example highlights perhaps one of the limitations to TGGLines and *TGGLinesPlus*: they depend on creating an image binary and then an image skeleton of the original input image, and so represents a thinned version of the original. In this way, for images like Figure 7, *TGGLinesPlus* will find a 1-pixel wide representation of the input image. This means that the *TGGLinesPlus* algorithm will create output paths that follow the interior of an input image's lines based on its skeleton representation. In contrast, PPHT and LSD, while presenting serious discontinuities between paths and presenting many small segments, perhaps non-intuitive paths, do a much better job of tracing the outlines of the original image itself.

EDLines does not generate any output for Image#10 in Figure 7. For the PPHT result in Figure 7, we initially thought it was an error with the plotting method, but we used the same plotting method for each image and the rest of the images are fine. So there is some source of noise or something in the image that the PPHT cannot handle. To add to this discussion, the original paper (Matas et al., 2000) seems to work fine on what the authors call "real" images. Since PPHT is gradient based (see Section 2) and the text recognition image has a lot of "shade" or "shadow" to it, which might affect the results. The result also might be affected by the fact that all other algorithms have many parameters that need tuning to get the best results, but that is outside the scope of this paper. Our *TGGLinesPlus* algorithm works well without tuning (see Table 4 in Section B.2 in the supplementary materials (see **DATA AND CODES AVAILABILITY STATEMENT**). Benchmark implementation specifications for each of the benchmark methods, including our *TGGLinesPlus*, are detailed in Section B.2 in the supplementary materials (see **DATA AND CODES AVAILABILITY STATEMENT**).

| Img # | Input images | Ground truth | PPHT | LSD | EDLines | Linelet | **TGGLines** | ***TGGLinesPlus*** |
|---|---|---|---|---|---|---|---|---|
| 01 | | | | | | | | |
| 02 | | | | | | | | |
| 03 | | | | | | | | |

Figure 4. Benchmark results part 1: small size (character) images: MNIST, CMINIST). Note that for readability, the results are cropped to the plain white/black background.

| Img # | Input images | Ground truth | PPHT | LSD | EDLines | Linelet | **TGGLines** | ***TGGLinesPlus*** |
|---|---|---|---|---|---|---|---|---|
| 04 | | | | | | | | |
| 05 | | | | | | | | |
| 06 | | | | | | | | |

Figure 5. Benchmark results part 2: binarized image input: road, retina, and cement cracks (i.e., concrete cracks).

| Img # | Input images | Ground truth | PPHT | LSD | EDLines | Linelet | **TGGLines** | ***TGGLinesPlus*** |
|---|---|---|---|---|---|---|---|---|
| 07 | | | | | | | | |
| 08 | | | | | | | | |
| 09 | | | | | | | | |

Figure 6. Benchmark results part 3: more complex images: Remote sensing Landsat 8 OLI imagery and contours image. Note that image #09 needs to be inverted (#07 and #08 are not inverted) first before feeding them into each algorithm, because the black part in the image is meaningful for line extraction.

| Input image *(Img #: 10)* | Ground truth | PPHT |
|---|---|---|
| | | |
| LSD | EDLines | Linelet |
| | | |
| **TGGLines** | ***TGGLinesPlus*** | |
| | | |

Figure 7. Benchmark results part 4 (for Img #10): a more complex example: documents. Our *TGGLinesPlus* outperforms all other methods, including TGGLines. For example, our *TGGLinesPlus* can segment all letters in “Segmentation” (i.e., the last word in the first line of the input image), except for the letter “i”. It segments “o” as a whole full loop, it segments “s” and the two “n”s as one line segment correctly, and it segments “m” and both “t” correctly: m to three line segments, because it has one junction node. Our algorithm segments the first “t” into four line segments, that is because it identifies one junction node and four terminal nodes. It segments the second “t” to three line segments, because our algorithm identifies one junction and three terminal nodes. Perhaps due to the scanning processes and low resolution, the skeleton algorithm missed the small mark on the left of the second letter “t”. Note that the EDLines output is blank, that means the algorithm cannot handle such types of images at all.

## 4.3 Quantitative evaluation

### 4.3.1 Ground truth

The general rules we used when we manually annotated the ground truth to benchmark are as follows: **(1)** The high level general guidance for line annotation judgment: create the line version that best represents the original benchmark image. **(2)** Keep the annotated lines connected/isolated if they are connected/isolated in the original image; thus where to end one line segment depends on how it meets with other lines it connects to. **(3)** Draw lines in the center of the image (like driving in the middle of a lane on the road). Human drawing benchmark Ground truth involves inevitable human errors shown as examples in Figure 8 below.

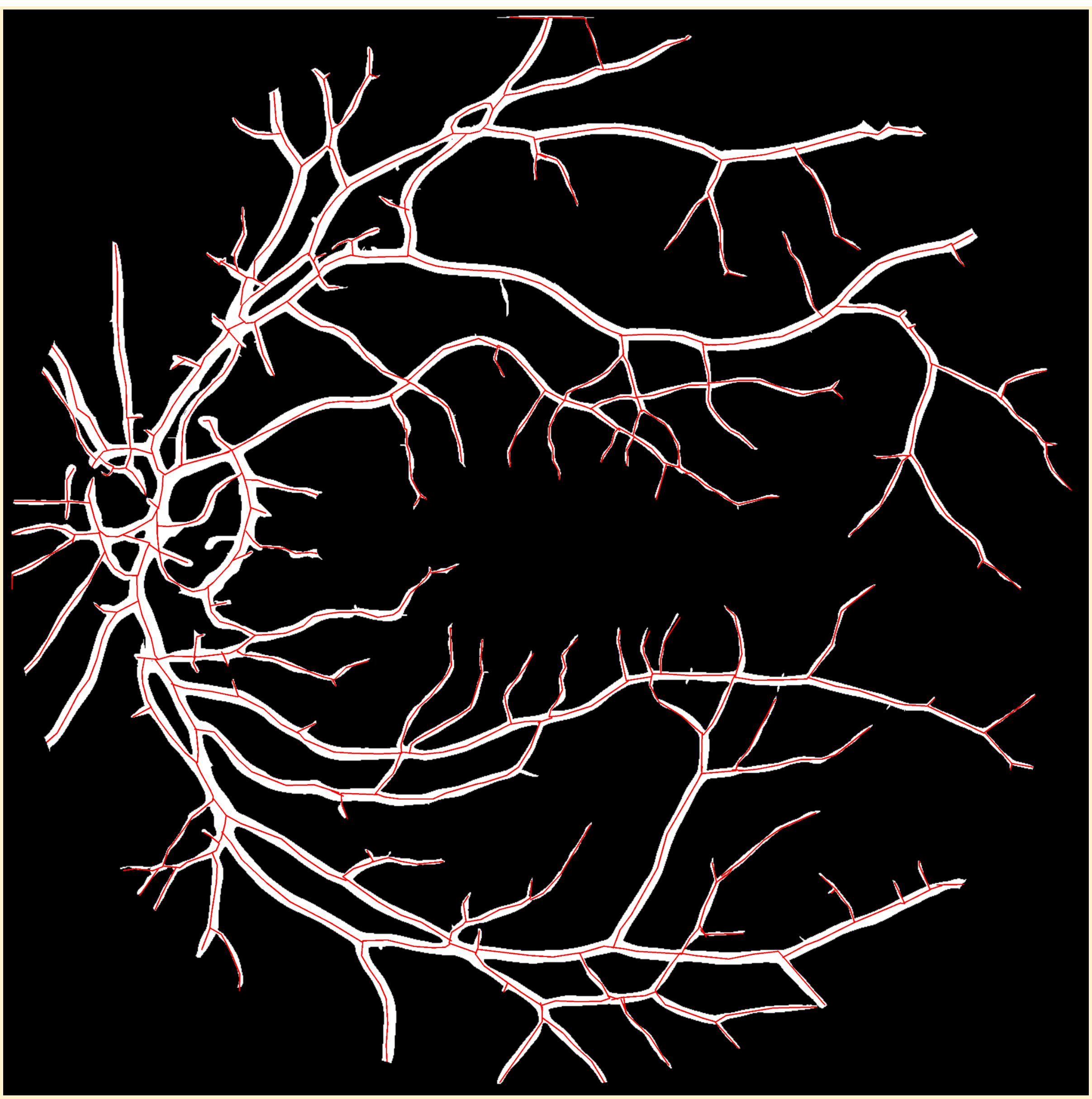

Figure 8. The ground truth, or human annotated, lines for the retina image #05 is an example of inevitable human errors when creating the ground truth drawing data. Line segments (perceived) are annotated in red by humans. This provides an example of how challenging it is for humans to annotate the lines representing the retina from medical imagery. It may look like there is only one connected component from the annotation, but human errors and hardware (e.g. mouse sensitivity) can make unexpected mistakes such as multiple clicks at the same position or clicks in undesired locations. This could be the reason why the connected components for image #5 are so high (see Table 1). In this scenario, TGGLines and *TGGLinesPlus* do an even better job than human annotation (the GT).

#### 4.3.2 Quantitative evaluation metrics

Quantitative evaluation metrics presented in this paper include calculating the total number of connected components, total number of junction nodes, and total number of terminal nodes. The evaluations were run as experiments on the human annotated GT drawings and the six line detection methods benchmarked. We chose these metrics through experimenting with different evaluation metrics (such as the total number of detected line segments, the total length of detected line segments, and total number of turning nodes). Our trial experiments show the three metrics (connected components, junction nodes, and terminal nodes) can be helpful for quantifiable evaluation when combined with the qualitative evaluation (i.e., visual checks, see Section 4.2).

#### 4.3.3 Quantitative evaluation for the Benchmark results

This section presents a table listing the evaluation results using the three quantitative evaluation metrics (Section 4.3.2) on the benchmark methods for the benchmark images. Although the example in Figure 8 shows that human annotations of lines have inevitable errors, the GT drawings provide something to compare and benchmark the images against.

Compared to the GT lines, the *TGGLinesPlus* performed the best, with TGGLines also performing similarly to the GT lines, with the exception of image #10 (the document page). Of the methods evaluated, the GT, LSD, TGGLines, and *TGGLinesPlus* were the approaches that were able to create an output for each image; however just because an output was created they were not all well replicated. EDLines were unable to create outputs for the character and digit images (The MNIST and the document page). The PPHT was unable to create reliable images for the MNIST and document images, even though it produced an output. LSD created an output for the MNIST images #01 to 03 but this method did not locate and replicate any junction nodes due to all line segments detected being not connected. Linelet performed similarly to LSD on the MNIST images, creating no output for image #01 and line outputs without junction nodes for images #02 and 03. For image #10, (the document page) the only line output that produced the number of connected components, junction nodes, and terminal nodes similar to the GT lines was the *TGGLinesPlus* approach.

In the literature reviewed for this paper (see Section 1 and Section 2), we noticed that quantitative metrics for evaluation were not often used and qualitative visual check evaluation was common. This may be because of issues with GT lines not truly being the ground truth, but instead a representation based on human visualization and computer interfaces. Although the quantitative evaluation metrics are imperfect, this paper seeks to be a basis on which other quantitative metrics for line detection evaluation can be measured. We believe that quantitative metrics (e.g., connected components and terminal node numbers per line output) can be utilized in conjunction with the qualitative visual examination to understand the robustness of the evaluated methods. See Tables 1 to 3 below for the results of the three quantitative metrics (connected components, junction nodes, and terminal nodes) found to be informative when coupled with qualitative evaluation.

Table 1. Quantitative evaluation metric: the total number of *connected components*.

| Img # | GT | PPHT | LSD | EDLines | Linelet | TGGLines | *TGGLinesPlus* |
|---|---|---|---|---|---|---|---|
| 01 | 1 | 5 | 8 | *NO OUTPUT* | *NO OUTPUT* | 1 | 1 |
| 02 | 4 | *NO OUTPUT* | 8 | *NO OUTPUT* | 4 | 4 | 4 |
| 03 | 8 | *NO OUTPUT* | 8 | *NO OUTPUT* | 3 | 4 | 3 |
| 04 | 9 | 21 | 6 | 31 | 55 | 6 | 6 |
| 05 | 12 | 87 | 11 | 159 | 403 | 1 | 1 |
| 06 | 3 | 122 | 114 | 2 | 82 | 2 | 2 |
| 07 | 43 | 60 | 119 | 33 | 63 | 14 | 49 |
| 08 | 80 | 117 | 189 | 47 | 110 | 30 | 81 |
| 09 | 59 | 443 | 370 | 84 | 396 | 13 | 22 |
| 10 | 283 | 10 | 87 | *NO OUTPUT* | 61 | 29 | 229 |

Note: *GT: ground truth. NO OUTPUT: refers to the corresponding method was not able to produce an output for the corresponding benchmark image (i.e., the output result was an empty image).*

Table 2. Quantitative evaluation metric: the total *number of junction nodes*.

| Img # | GT | PPHT | LSD | EDLines | Linelet | TGGLines | *TGGLinesPlus* |
|---|---|---|---|---|---|---|---|
| 01 | 4 | 11 | 0 | *NO OUTPUT* | *NO OUTPUT* | 5 | 7 |
| 02 | 6 | *NO OUTPUT* | 0 | *NO OUTPUT* | 0 | 3 | 1 |
| 03 | 13 | *NO OUTPUT* | 0 | *NO OUTPUT* | 0 | 4 | 6 |
| 04 | 234 | 365 | 541 | 246 | 942 | 190 | 140 |
| 05 | 373 | 1241 | 1428 | 332 | 923 | 328 | 285 |
| 06 | 25 | 138 | 187 | 12 | 16 | 51 | 80 |
| 07 | 103 | 534 | 190 | 458 | 806 | 88 | 52 |
| 08 | 146 | 2107 | 423 | 1354 | 1572 | 257 | 131 |
| 09 | 664 | 2454 | 1957 | 1360 | 2115 | 796 | 920 |
| 10 | 432 | 1224 | 23 | *NO OUTPUT* | 6 | 10 | 253 |

Note: *GT: ground truth. NO OUTPUT: refers to the corresponding method was not able to produce an output for the corresponding benchmark image (i.e., the output result was an empty image).*

Table 3. Quantitative evaluation metric: the total *number of terminal nodes.*

| **Img #** | **GT** | **PPHT** | **LSD** | **EDLines** | **Linelet** | **TGGLines** | ***TGGLinesPlus*** |
|---|---|---|---|---|---|---|---|
| 01 | 0 | 17 | 16 | *NO OUTPUT* | *NO OUTPUT* | 3 | 1 |
| 02 | 10 | *NO OUTPUT* | 16 | *NO OUTPUT* | 8 | 9 | 9 |
| 03 | 22 | *NO OUTPUT* | 14 | *NO OUTPUT* | 6 | 12 | 12 |
| 04 | 114 | 120 | 120 | 124 | 263 | 73 | 74 |
| 05 | 220 | 414 | 372 | 370 | 982 | 155 | 207 |
| 06 | 11 | 256 | 275 | 6 | 158 | 27 | 74 |
| 07 | 123 | 320 | 325 | 70 | 441 | 69 | 136 |
| 08 | 228 | 837 | 527 | 116 | 670 | 179 | 239 |
| 09 | 172 | 1615 | 1359 | 412 | 1194 | 152 | 277 |
| 10 | 603 | 210 | 179 | *NO OUTPUT* | 122 | 58 | 441 |

Note: *GT: ground truth. NO OUTPUT: refers to the corresponding method was not able to produce an output for the corresponding benchmark image (i.e., the output result was an empty image).*

## 5. DISCUSSION, STRENGTHS, LIMITATIONS, AND FUTURE WORK

### 5.1 Discussion

Our *TGGLinesPlus* algorithm is meant to be intuitive and straightforward for researchers and practitioners from a wide and diverse range of disciplines such as CV, ML, graph theory, image processing, remote sensing, and the art and design communities. Our method produces well-segmented graph paths based on primary junction and terminal nodes (see Section 3.3). Our implementation of *TGGLinesPlus* is open source, robust, and modular. While designing the algorithm, we prioritized having a method and codebase with easily-understood, well-documented methods that could be individually improved without affecting the overall algorithm pipeline. Additionally, we wanted to create user-friendly methods that return useful printouts at different stages of the algorithm. We use NetworkX (Hagberg et al., 2008) to implement graphs, as it is the most comprehensive graph theory library in Python and it is optimized and well-documented. *TGGLinesPlus* works with undirected graphs, thus the path direction does not matter.

Existing relevant methods (e.g., TGGLines (Gong et al., 2020), skan (Nunez-Iglesias, 2016)) include turning nodes in their graph node types. In our *TGGLinesPlus*, we remove turning nodes to segment the paths, using only the terminal nodes and primary junction nodes to segment paths (detailed in Section 3.3). We do this because turning nodes bring in some unnecessary details and thus unnecessary complexity in the pixel space. For some cases TGGLines works worse than *TGGLinesPlus* (See image #01, #02, #03 in Figure 4 and image #10 in Figure 7). The poor performance of TGGLines in those images could be partially caused by the tiny trivial paths segmented by "artifact" turning nodes. In addition, TGGLines utilized the Douglas-Peucker algorithm (Douglas & Peucker, 1973) to simplify each segmented path, though we do not do this in *TGGLinesPlus* because we do not use turning nodes to segment our paths. Thus, the simplification part in TGGLines would not work well for *TGGLinesPlus*. Also the needs of further simplifying segmented paths using turning nodes depend on users, as different domains have different needs. For example, in terms of spatial resolution in remote sensing imagery, a tiny detail of 2-3 pixels can be mighty in a low resolution satellite image. Our output includes a dictionary of useful information (detailed in Algorithm 1 in Section 3.5 and also in the step 5 in Section A in the supplementary materials (see **DATA AND CODES AVAILABILITY STATEMENT**), users can perform some post-processes if they need to break segmented paths based on turns.

In our benchmark experiments and results (Section 4.2), we did not include machine /deep learning-based line segmentation methods because it is not fair/relevant. Our *TGGLinesPlus* does not require any training data and/or feature engineering, which are required for almost all supervised ML and DL algorithms. We provide the machine intelligence community a different option to complete the task of line segment detections. We believe that simplicity, robustness, and less labor-intensiveness should be prioritized when people choose a method for their work. Simplicity is a rule of thumb (Blumer et al., 1987; Domingos, 1999; Paola & Leeder, 2011).

### 5.2 Strengths

In this section, we discuss how our *TGGLinesPlus* differs from the benchmark methods we introduced in Section 2, and Section 4.2 and Section 4.3. Table 4 below provides an overview of strengths. The strengths are not ordered by importance, but more relevant strengths are nearby each other.

Table 4. An overview of strengths

| ***Strengths*** | **PPHT** | **LSD** | **EDLines** | **Linelet** | **TGGLines** | ***TGGLinesPlus*** |
|---|---|---|---|---|---|---|

| Self-contained | × | × | × | × | × | ✓ |
|---|---|---|---|---|---|---|
| Open-source | ✓ | ✓ | ✓ | ✓̶ | × | ✓ |
| Detecting junction nodes and primary junction nodes | × | × | × | × | ✓̶ | ✓ |
| Detecting self-looped lines | × | × | × | × | × | ✓ |
| Detecting turning nodes | × | × | × | × | ✓ | × |
| Remaining topology | ✓̶ | ✓̶ | × | ✓̶ | ✓̶ | ✓ |
| Robustness | × | ✓̶ | × | ✓̶ | ✓̶ | ✓ |
| High usability and utility of the output results | × | × | × | × | × | ✓ |
| Potential for many real-world applications | × | × | × | × | ✓̶ | ✓ |
| Avoids human error but remains reliable when compared to GT | × | ✓̶ | × | × | ✓̶ | ✓ |

**Note:** OCR refers to optical character recognition. ✓ indicates a corresponding strength is present in a corresponding method, × refers to a method that does not reflect a corresponding strength. ✓̶ indicates a corresponding strength is not fully or reliably present.

1) ***Self-contained*** *and thus reproducible and replicable.* As we discussed in Sarigai et al. (2024), Python-based implementation is not easy to keep fully self-contained and thus not always easy for reproduction and replication (it is always necessary to check the used Python libraries version, and dependency etc). However, we make the reproducibility of our algorithm and code transparent and convenient through the following: we provide a YAML file (.yml), which is a Conda environment description of relevant dependencies and is provided in our GitHub repository (see **DATA AND CODES AVAILABILITY STATEMENT**). In addition, we also provided code in our GitHub repository for how to save output results as a file ( .pkl file) to allow users to do further analysis of the output results, instead of needing to rerun the algorithm every time.
2) ***Open source:*** The implementation of our *TGGLinesPlus* is fully open source, and fully based on open-source libraries. This furthers the goals of software democratization. Code (both jupyter notebook files and the main algorithm code Python script) includes clear and explanatory programming comments that allow users to informatively modify and customize the code for their needs. The code for Linelet is indeed provided in their published article. However it is implemented in Matlab, which is not an open source software/platform, that is the reason we classified it as half open source.

3) ***Detecting junction nodes and primary junction nodes:*** TGGLines detects junction nodes, but does not cover the detection of primary junction nodes. *TGGLinePlus* output include different levels of the output results, connected component, segmented path (line segments), detected junction nodes, and primary junction nodes.
4) ***Detecting self-looped lines:*** no benchmarked methods, including our previous TGGLines could detect self-looped lines, our *TGGLinesPlus* can. Self-looped lines refers to a curved line that starts and ends with the same terminal node or junction node. See image #4 (in Figure 5) and image #10 (in Figure 7) for example (note, need to zoom in to see the self-loop detail in image #4 in Figure 5).
5) ***Detecting turning nodes:*** in all methods, only TGGLines detected tuning nodes. In our *TGGLinesPlus*, we purposely do not detect turning nodes, because turning nodes from topological graphs constructed from images do not always accurately reflect its topology due the fact that a pixel has 8 direct neighboring pixels. Also, if some applications need to detect turning nodes, the comprehensive result output of our *TGGLinesPlus* can help them meet that need through some further analysis.
6) ***Remaining topology:*** (i.e., shape and connectivity) of objects/features in the original images. From Section 4.2, we can visually check that *TGGLinesPlus* can preserve the topology in the original images very well.
7) ***Robustness:*** *TGGLinesPlus* is robust in terms of the ability to detect line segments from a wide range of imagery types and domains.
8) ***High usability and utility of the output results:*** The output of our *TGGLinesPlus* algorithm is easily reusable for further analysis, and contains comprehensive information. The output for *TGGLinesPlus* can be in either vector or raster format (as raster in any resolution can always be generated from a corresponding vector format). Exemplary output dictionary of our *TGGLinesPlus* is provided and detailed in Section 3.4.3.
9) ***Potential for many real-world applications:*** *TGGLinesPlus* is very promising to advance many real-word applications (detailed in Section 5.4), including geospatial applications (Section 5.4.1) and more general applications in computer graphics, image understanding, and computer science (Section 5.4.2).
10) ***Avoids human error but remains reliable when compared to GT:*** We gave a full mark for *TGGLinesPlus*, because it is robust across a wide range of images (through both qualitative and quantitative evaluation, Section 4.2 and Section 4.3), and TGGLines and LSD a half mark, because TGGLines performs pretty well for most images expect for image #1 to image 3, and LSD can perform fairly well, besides an issue with often detecting double lines for each single line representation in the original image.
11) ***Additional informative and self-contained guide materials for users.*** To aid in users' experience when developing their own needs from our *TGGLinesPlus* algorithm and Python implementation, we provide users the following useful and easily accessible additional supplementary materials for understanding the *TGGLinesPlus* algorithm in the GitHub repository (see **DATA AND CODES AVAILABILITY STATEMENT**). More specifically, the supplementary materials are about the computing environment for benchmark and a more in-depth discussion about our Python implementation of *TGGLinesPlus*.

## 5.3 Limitations and future work

*TGGLinesPlus* works pretty well on various images from different domains, though it has some limitations and meanwhile indicates potential future work directions. One major limitation is that it relies on image binary and skeleton results. More specifically, if binary and skeleton algorithms are not robust enough to keep the connectivity of the original image, our *TGGLinePlus* will not be able to detect lines appropriately. See Section A *Step 0* in the supplementary materials (see **DATA AND CODES AVAILABILITY STATEMENT**) for how different algorithms of binarization and skeletonization would affect the connectivity of an original image. Thus, robust and adaptive binarization and skeleton algorithms would make our algorithm work even better. Noisy skeletons or images with other objects in the image will become part of the skeleton, meaning that our method will not work appropriately. For example, let us revisit the example in Figure 6 image #09 (the contours). Even with a high-resolution image, the skeletonization process introduces many artifacts (i.e., tiny junctions) that are not present in the original image. Therefore, image resolution is perhaps just as important as the binarization and/or skeletonization methods to the final result of our algorithm. In essence, the runtime of *TGGLinesPlus* is significantly impacted by the artificial introduction of junctions and is unrelated to the algorithm itself. In addition, for now, our method works only for 2D images, not 3D images. In the future, we would like to extend *TGGLinesPlus* towards 3D imagery and also to apply our algorithm to solve real-world problems in more application domains.

Our method is currently fast enough to run on a desktop computer, where complex cases take less than 1 minute to complete. In terms of further performance improvements, though, the most time-intensive part of our implementation is path segmentation (Algorithm 3 in Section 3.5). *TGGLinesPlus* leverages NetworkX (Hagberg et al., 2008) to create skeleton graphs, which is relatively quick, but our algorithm uses a sequential search to iterate over lists of path segmentation endpoints to find unique paths in a graph. It also creates a temporary copy of a given subgraph to keep track of which nodes have already been visited in the path segmentation process by deleting those nodes and removing them from the path search (for more information on this point, see the Step 3.2.2 in Section A in the supplementary materials (see **DATA AND CODES AVAILABILITY STATEMENT**). What is more, we filter important topological information about each subgraph at different stages of the algorithm using Python list comprehensions. A significant speedup of the algorithm could come from: (1) finding a non-sequential path search algorithm, (2) eliminating the need to create subgraph copies to track visited nodes, or (3) keeping track of topological information and unique paths data without the need to filter them later. In addition, *TGGLinesPlus* algorithm is currently structured as the following: processing each subgraph sequentially from the main whole graph, which significantly speeds up processing since the search through all nodes in a large graph is expensive. The subgraph structure makes it possible to parallelize computing multiple subgraphs simultaneously in the future. This parallelization upgrade could be very important for analyzing very large and high-resolution datasets to further improve *TGGLinesPlus* performance.

## 5.4 Potential applications

As we briefly provided some potential strengths in Section 5.2 above. Here we provide some more specific potential applications of our proposed *TGGLinesPlus* algorithm, in terms of some geospatial problems setting and also in a more general context.

### 5.4.1 Potential geospatial applications

Geospatial applications of *TGGLinesPlus* include but are not limited to the following: **(1) Automating digitization task in geographic information systems (GIS):** Digitization is a classic and important task in GIS. In the old days, digitization might be one of the most labor intensive and costly steps in a GIS project lifecycle. Our *TGGLinesPlus* has a great potential to advance and automate the cumbersome task of digitization in GIS (see image #4 in Figure 5, and images #7 to #9 in Figure 6). **(2) Automating water resources monitoring from remote sensing data:** Earth observation data such as remote sensing imagery plays a major role in monitoring the planet. Our *TGGLinesPlus* is very promising to automate the water resources such as river networks from remote sensing imagery. More specifically, river data can be extracted from remote sensing imagery (including different resolution, such Landsat and Sentinel) using existing index such as Normalized Difference Water Index (NDWI) and Modified Normalized Difference Water Index (MNDWI), then can use our *TGGLinesPlus* to automate the extraction and vectorization of river networks, including rivers downstream (see image # 5 (retina) in Figure 5 which is very similar to river networks in a geospatial setting). **(3) Curved line detection for contour line map digitization:** Contour line maps often have many small complex line segments making up a larger topographic image. *TGGLinesPlus* has the ability to detect curved lines, which uniquely situates it to digitize contour line imagery (see image # 09 in Figure 6). **(4) Digitizing text from maps:** *TGGLinesPlus* has strong OCR capabilities allowing this approach to reliably digitize characters from scanned imagery. Implementing the *TGGLinesPlus* approach can both digitize the geographic content of a map and the textual labels, map elements and marginal information (see image #10 in Figure 7). **(5) Digitizing lines to be brought from a GIS into a design software:** GIS software is often used for data analysis by professionals and cartographers. Professionals may take GIS outputs and bring them into design software to be polished before a final map is produced. *TGGLinesPlus* can assist in bringing various vector lines into a design software when the GIS output is not reliably able to be parsed in the design software.

### 5.4.2 Potential more general applications

Beyond the geospatial applications we detailed in Section 5.4.1 above, our *TGGLinesPlus* has great potential for the advancement of many applications in image understanding, computer graphics, and computer science. Those include but not limited to the following: **(1) Image into editable text:** OCR capabilities of *TGGLinesPlus* allow scanned images to be turned into editable text (see Image #10 in Figure 7). **(2) Advancing medical images analysis:** Medical images such as computed tomography (CT) scan and ultrasound imaging (see Image # 5 in Figure 5). **(3) Advancing image retrieval:** image retrieval, especially retrieval of scientific images remains very challenging see Yang et al. (2020), including patent images such as engineering drawing. The line segments detected by our *TGGLinesPlus* could keep the topology of objects in original images and retrieve vector objects would be promising to improve both image retrieval performance and accuracy. **(4) Computational geometry analysis:** from images and hand drawn diagrams. *TGGLinesPlus* can assist in geometrical analysis. The image and vector output of *TGGLinesPlus* can make computation geometry analysis more efficient. **(5) Computer graphics design:** for example in art work: *TGGLinesPlus* can help convert hand drawn designs into digital formats. In addition, *TGGLinesPlus* can support creating very high resolution image outputs. Any vector output can be rasterized into a very high resolution to be used by artists. **(6) Image vectorization:** Image vectorization is the process of converting a raster image into a corresponding vector image that represents the same content (e.g. objects) in the original raster image. Image vectorization is still computationally challenging, as discussed in a recent review (Dziuba et al., 2024). *TGGLinesPlus* can assist in restyling of objects from images, such as selecting a circle in the vectorized image.

## 6. CONCLUSION

In this paper, we propose *TGGLinesPlus*, a topological graph-guided algorithm for line extraction and vectorization from binary images. The main motivation is the need to design and implement a robust and intuitive algorithm for line feature extraction from grayscale images. Our experiments, including benchmarking five state-of-the-art line detection methods, demonstrates the robustness and wide applicability of *TGGLinesPlus*. On various domain images (small to large sizes, simple to complex scenarios, document/character to contours and remote sensing imagery), *TGGLinesPlus* competitively outperforms other methods. We have not compared our method with machine /deep learning methods, because it is not a fair comparison. (1) *TGGLinesPlus* does not require any training data as many supervised machine/deep learning methods do. (2) *TGGLinesPlus* does not require any parameters and thus no need to tune parameters, whereas machine/deep learning methods often have many parameters and require parameter tuning to get best results. Many important problems can be moved forward through *TGGLinesPlus*, including but not limited to the following: advanced optical character recognition (OCR) techniques (it is very promising, see the results of document recognition in Figure 7), road lane line extraction for real-time autonomous driving, concrete cracks analysis in structural engineering, AutoCAD map vectorization, contour map digitalization, medical image processing and analysis, in addition to feature extraction for machine learning algorithms.

## DATA AND CODES AVAILABILITY STATEMENT

The implementation of our *TGGLinesPlus* algorithm is in Python with many examples from the paper included in Jupyter Notebooks. The code is freely available to the public and can be accessed at the GitHub repository: https://github.com/GeoAIR-lab/TGGLinesPlus. More specifically, see below.

- The YAML file "environment.yml" can be found at the root of the *TGGLinesPlus* GitHub Repository. In the README file of the repository, we have included a very informative instruction for how to quickly and easily create the computing environment to implement our *TGGLinesPlus* using the thoughtful .yml file we provided.
- The Jupyter Notebook file for the full exemplary output for our *TGGLinesPlus* algorithm using the algorithm illustration example (Figure 2 and Figure 3 in the paper above) can be found in the folder at this link (https://github.com/GeoAIR-lab/TGGLinesPlus/tree/main/notebooks)
    - This Jupyter Notebook file also includes code snippets for how to save output results as a file (.pkl file) and later load the saved .pkl file back to allow users to do further analysis of the output results, instead of needing to rerun the algorithm every time. For high-resolution and complex images such as remote sensing imagery, this thoughtful technical tip can be a game changer for many domain experts who may not have the technical background to think of this elegant solution.

- The Jupyter Notebook file for the benchmarked domain examples of our *TGGLinesPlus* algorithm can be found in the folder at this link (https://github.com/GeoAIR-lab/TGGLinesPlus/tree/main/notebooks)
- Supplementary materials for the *TGGLinesPlus* algorithm implementation (see the PDF file in this folder at this link: https://github.com/GeoAIR-lab/TGGLinesPlus/tree/main/supplementary_materials_TGGLinesPlus_paper)

## CONFLICT OF INTEREST STATEMENT

No conflict of interest was reported by the authors.

## ACKNOWLEDGMENTS

This work was supported in part by a grant from the US National Aeronautics and Space Administration under Grant number 80NSSC22K0384 and by the funding support from the College of Arts and Sciences at University of New Mexico. The authors are very grateful to Professor Emeritus Alan M. MacEachren at Penn State for his very inspiring discussion about simplicity. The authors are also very grateful to the editors (Dr. John P. Wilson and Dr. Kevin M. Curtin) and the anonymous reviewers for their very helpful and constructive suggestions.